\def\paperTitle{EvoVid: Temporal-Centric Self-Evolution \\for Video Large Language Models}

\newcommand{\ours}{\textbf{EvoVid}\xspace}

\def\authorBlock{
\fontsize{11.5}{14}\selectfont
Shiqi Huang$^1$, Ziyue Wang$^2$, Zhongrong Zuo$^2$, Han Qiu$^2$, Qi She$^2$, Bihan Wen$^1$\thanks{Corresponding Authors}\\[4pt]

\fontsize{10.5}{14}\selectfont
$^1$School of Electrical and Electronic Engineering, Nanyang Technological University\\
$^2$ByteDance
}

\newif\ifreview 
\newif\ifarxiv \newcommand{\arxiv}{\arxivtrue}
\newif\ifcamera 

\arxiv

\pdfoutput=1
\documentclass{article}

\PassOptionsToPackage{numbers, compress}{natbib}
 
\ifreview \usepackage{meta/neurips_2026} \fi
\ifarxiv \usepackage[preprint]{meta/neurips_2026} \fi
\ifcamera \usepackage[main, final]{meta/neurips_2026} \fi

\usepackage{fix-cm}
\usepackage{array}
\usepackage{bbm}
\usepackage{nicematrix}

\usepackage{tipa}
\usepackage{dsfont}
\usepackage{etoolbox}  

\usepackage[noend]{algorithmic}
\usepackage{algorithm}

\usepackage{float}
\usepackage{newfloat}
\usepackage{listings}
\floatstyle{ruled}
\newfloat{listing}{tb}{lst}{}
\floatname{listing}{\small Algorithm}

\definecolor{mygray}{RGB}{234,234,234}

\usepackage{adjustbox}

\definecolor{darkgreen}{rgb}{0.13, 0.55, 0.13}

\newcommand{\bgap}[1]{\rlap{\fontsize{7pt}{1em}\selectfont\,\textsuperscript{\textcolor{cyan}{\textbf{$\uparrow$#1}}}}}  

\usepackage{graphicx}	
\usepackage{amsmath}	
\usepackage{amssymb}	
\usepackage{booktabs}
\usepackage{times}
\usepackage{epsfig}
\usepackage{caption}
\usepackage{float}
\usepackage{placeins}
\usepackage{color, colortbl}
\usepackage{stfloats}
\usepackage{enumitem}
\usepackage{tabularx}
\usepackage{xstring}
\usepackage{multirow}
\usepackage{xspace}
\usepackage{url}
\usepackage{subcaption}

\ifcamera \usepackage[accsupp]{axessibility} \fi

% ...

\ifarxiv  \fi

\newcommand{\R}[1]{{%
    \textbf{%
        \ifstrequal{#1}{1}{\textcolor{red}{R#1}}{%
        \ifstrequal{#1}{2}{\textcolor{blue}{R#1}}{%
        \ifstrequal{#1}{3}{\textcolor{magenta}{R#1}}{%
        \ifstrequal{#1}{4}{\textcolor{teal}{R#1}}{%
                           \textcolor{cyan}{R#1}%
        }}}}%
    }%
}}

\definecolor{Gray}{gray}{0.5}
\definecolor{nicergreen}{rgb}{0.13, 0.54, 0.13}
\definecolor{nicered}{rgb}{0.83, 0.16, 0.16}
\definecolor{lightgray}{RGB}{230, 230, 230}
\definecolor{Highlight}{HTML}{39b54a}  %

\usepackage{appendix}
\usepackage{footnote}

\usepackage{microtype}
\usepackage{cuted}
\usepackage{tocloft}

\usepackage[T1]{fontenc}
\usepackage{DejaVuSans}

\usepackage{ifsym, marvosym}

\let\svthefootnote\thefootnote
\newcommand\freefootnote[1]{%
  \let\thefootnote\relax%
  \footnotetext{#1}%
  \let\thefootnote\svthefootnote%
}

\makeatletter
\DeclareRobustCommand\onedot{\futurelet\@let@token\@onedot}
\def\@onedot{\ifx\@let@token.\else.\null\fi\xspace}

\def\eg{\emph{e.g}\onedot} 
\def\ie{\emph{i.e}\onedot}

\makeatother

\usepackage{amsthm}
\theoremstyle{plain}

\usepackage[framemethod=default]{mdframed}

\usepackage{wrapfig}
\usepackage{tcolorbox}
\tcbuselibrary{listings, skins, breakable}  

\usepackage{xr-hyper}

\makeatletter
\newcommand*{\addFileDependency}[1]{
  \typeout{(#1)}
  \@addtofilelist{#1}
  \IfFileExists{#1}{}{\typeout{No file #1.}}
}

\makeatother

\definecolor{cvprblue}{rgb}{0.21,0.49,0.74}
\usepackage[pagebackref,breaklinks,colorlinks]{hyperref}
\usepackage[capitalize]{cleveref}
\crefname{section}{Sec.}{Secs.}
\crefname{table}{Table}{Tables}
\crefname{figure}{Fig.}{Figs.}

\usepackage[utf8]{inputenc} 
\usepackage[T1]{fontenc}    
\usepackage{url}            
\usepackage{booktabs}       
\usepackage{amsfonts}       
\usepackage{nicefrac}       
\usepackage{microtype}      
\usepackage{xcolor}         

\usepackage{amssymb}  
\usepackage{colortbl}  
\newcommand{\cmark}{\checkmark}

\newcommand{\xmarkg}{\textcolor{gray}{\texttimes}}
\usepackage{bm}
\usepackage{bbm}
\usepackage{fontawesome5}
\begin{document}
\title{\paperTitle}
\author{\authorBlock}
\maketitle

\begin{abstract}
Recent Video Large Language Models (Video-LLMs) have demonstrated strong capabilities in video reasoning through reinforcement learning (RL). However, existing RL pipelines rely heavily on human-annotated tasks and solutions, making them costly to scale and fundamentally constrained by human expertise. Self-evolving frameworks have recently emerged as a promising alternative through autonomous Questioner–Solver self-play. Unfortunately, these approaches are primarily designed for static modalities such as text and images, fundamentally failing to capture the temporal dynamics that are central to video reasoning. In this work, we propose \ours, a temporal-centric self-evolving framework that enables Video-LLMs to improve directly from raw, unannotated videos. Specifically, we introduce two complementary temporal-centric rewards: a temporal-aware Questioner reward that encourages temporally dependent question generation through temporal perturbation sensitivity, and a temporal-grounded Solver reward that provides automatic temporal supervision via inherent video segment localization. Extensive experiments across four base models and six benchmarks demonstrate consistent improvements over both base models and existing self-evolving baselines, achieving competitive performance with supervised methods. These results highlight temporal-centric self-evolution as an effective and scalable paradigm for video understanding and reasoning.
\end{abstract}

\section{Introduction}
\label{sec:intro}

Recent Video Large Language Models (Video-LLMs) have demonstrated strong capabilities in video understanding and reasoning, often enabled by reinforcement learning (RL)~\cite{feng2025video,li2025videochat,park2025deepvideo,wang2025time,wang2026video}. Despite this progress, these advancements still rely on human-annotated tasks and solutions to construct verifiable reward signals, making them costly to scale and inherently bounded by human expertise~\cite{zhao2025absolute, huang2025r}. This bottleneck is especially acute in the video domain, where the intricate interplay of spatial variance and temporal causality makes exhaustive, high-quality annotation exceptionally difficult~\cite{xu2019self}. 

\begin{figure}[t]
    \centering
    \includegraphics[width=0.95\linewidth]{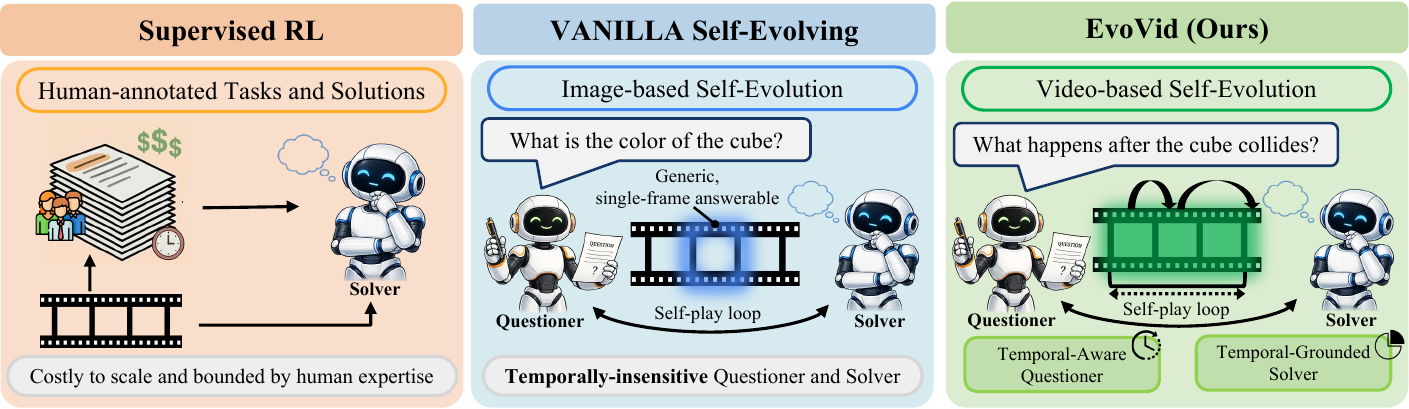}
    \caption{\textbf{Comparison between supervised RL, \textsc{vanilla} self-evolving frameworks, and \ours{}.} \textbf{Left:} Supervised RL relies on human-annotated tasks and solutions to construct reward signals, making training costly and inherently bounded by human expertise. \textbf{Middle:} \textsc{vanilla} self-evolving frameworks, primarily designed for static modalities, \ie, images, generate generic and often single-frame answerable questions, leading to temporally insensitive Questioner–Solver self-play. \textbf{Right:} \ours{} introduces video-based self-evolution through a temporal-aware Questioner and a temporal-grounded Solver, enabling temporal-centric self-evolution directly from raw, unannotated videos.}
    \label{fig:teaser}
    \vspace{-1em}
\end{figure}

To bypass this reliance on external supervision, self-evolving models have recently emerged as a promising alternative. Instead of learning from predefined annotations, these frameworks improve through iterative Questioner–Solver self-play, where both training tasks and learning signals are generated inherently during the RL process~\cite{huang2025r, yang2025spell,he2025visplay, thawakar2025evolmm}. While the self-evolving paradigm has shown strong effectiveness in Large Language Models (LLMs)~\cite{zhao2025absolute, huang2025r, yang2025spell} and is gaining increasing attention in Multimodal Large Language Models (MLLMs)~\cite{he2025visplay, thawakar2025evolmm}, existing frameworks are primarily designed for static modalities such as text and images, failing to account for the temporal dynamics inherent to video. Consequently, directly extending these methods to video often results in temporally insensitive and modality-agnostic self-evolution. 

As shown in Figure \ref{fig:teaser}, existing \textsc{vanilla} self-evolving frameworks typically generate generic questions that can be answered from a single frame or static appearance cues alone. This limitation largely stems from the use of generic rewards, such as difficulty and diversity~\cite{he2025visplay}, which provide little incentive to capture temporal dependencies. Consequently, the generated questions remain largely temporally insensitive, and the resulting supervision fails to encourage reasoning over temporal order, causality, or state transitions. As a result, both the Questioner and Solver become insensitive to the temporal structure that is fundamental to video understanding. This reveals a key limitation of current self-evolving frameworks: they lack temporal-centric and verifiable learning signals that explicitly exploit the dynamic nature of video.

To address this limitation, we propose \ours, a temporal-centric self-evolving framework for video understanding and reasoning through a temporal-aware Questioner and a temporal-grounded Solver. Specifically, we design two complementary rewards targeting the Questioner and Solver respectively. First, we propose a temporal-aware Questioner reward that encourages the generation of temporally dependent questions. Our central insight is that meaningful video questions should rely on the correct temporal ordering of frames, making temporal sensitivity a natural intrinsic supervision signal. We therefore compare Solver responses on the original video and temporally perturbed variants, rewarding questions whose answers significantly degrade under temporal perturbation. This transforms temporal sensitivity into a label-free and verifiable learning signal, guiding the Questioner to generate temporally intensive queries.

Second, we introduce a temporal-grounded Solver reward that provides explicit temporal supervision without requiring human annotations. During question generation, the Questioner only observes a sampled temporal window from the video and generates questions conditioned on this partial context. The sampled window therefore naturally defines an inherent temporal segment associated with the generated question. The Solver is then trained to predict both the answer and the relevant temporal segment from the full video, while being rewarded for accurately localizing the sampled window. In this way, temporal grounding supervision emerges automatically from the self-evolving process itself, establishing grounded reasoning without requiring manually annotated timestamps.

Together, the temporal-aware Questioner and temporal-grounded Solver form a co-evolution process in which a single base model alternates between the two roles: the Questioner generates temporally dependent questions, while the Solver answers the questions and localizes the corresponding temporal segments. Both roles are jointly optimized using Group Relative Policy Optimization (GRPO)~\cite{shao2024deepseekmath}. By integrating temporal perturbation sensitivity and temporal grounding supervision into the self-evolving loop, \ours enables Video-LLMs to progressively acquire stronger temporal reasoning capabilities directly from raw, unannotated videos.

Our main contributions are three-fold:
\begin{itemize}
    \item We propose \ours, the first self-evolving framework for Video-LLMs which improves video understanding and reasoning directly from raw, unannotated videos without relying on human supervision.
    \item We introduce two temporal-centric and modality-inherent verifiable rewards: a temporal-aware Questioner reward that preserves temporal perturbation sensitivity, and a temporal-grounded Solver reward that leverages inherent temporal segment supervision.
    \item We conduct extensive experiments across four MLLMs and six video understanding and reasoning benchmarks, demonstrating consistent improvements and validating the effectiveness of video-based self-evolution.
\end{itemize}

\section{Related Work}
\label{sec:related}

\paragraph{Reinforcement Learning for Video-LLMs.}
Recent work has explored reinforcement learning (RL) for Video-LLMs through a range of video-specific optimization strategies. These include encouraging temporal sensitivity by perturbing frame order~\cite{feng2025video}, enhancing spatio-temporal perception via structured or rule-based reward fine-tuning~\cite{li2025videochat}, and stabilizing training with difficulty-aware reward formulations~\cite{park2025deepvideo}. Other approaches focus on temporally grounded tasks, such as leveraging IoU-based rewards for temporal video grounding with annotated segments~\cite{wang2025time}, or incorporating self-verification processes to assess search completeness in long-form video understanding~\cite{pan2025timesearch}. More recent efforts further investigate token-level credit assignment to better capture reasoning-relevant information during optimization~\cite{wang2026video}. Despite these advances, existing methods largely rely on human-labeled supervision such as annotated QA pairs, temporal segments, or task-specific heuristics to construct reward signals. This dependence limits scalability and constrains the model’s ability to generalize beyond the scope of human-provided annotations. In contrast, our approach derives verifiable reward signals directly from the temporal dynamics of raw video, enabling label-free optimization and supporting self-evolving reasoning without human supervision.

\paragraph{Self-evolving Models.}
Self-evolution~\cite{chen2401self, tao2024survey} has emerged as a powerful paradigm for eliciting reasoning behavior in Large Language Models (LLMs)~\cite{zhao2025absolute, kuba2025language, liu2025spice} without human labels, by utilizing intrinsic output consistency~\cite{huang2025r, yang2025spell} rather than external supervision. This paradigm has recently been extended to Vision-Language Models (VLMs)~\cite{li2025self}, typically instantiated as a questioner–solver framework, where both components co-evolve under internal verification signals such as voting-based aggregation~\cite{he2025visplay, wang2026v, wen2025self}, game outcomes~\cite{wang2025vision}, and continuous or trajectory-level supervision~\cite{thawakar2025evolmm, sunil2026ireasoner}. In parallel, some approaches exploit privileged 3D structure as a programmable verifier to enhance spatial reasoning~\cite{li2026mm}. However, these methods are primarily designed for text or static images, where verification signals lack inherent sensitivity to temporal structure. Consequently, directly extending them to video often leads to temporally agnostic question generation and limited performance gains. Our work addresses this limitation by introducing temporal-centric verifiable signals that are intrinsically tied to the time dimension of video data.
  
\section{Method}
\label{sec:method}

\subsection{Self-evolving Framework}
\label{sec:prelim}

\paragraph{Problem Formulation.}
In our work, the self-evolving model uses no question-answer annotations, metadata, or external reward models. The paradigm builds upon two roles initialized from a single pretrained base model: a \emph{Questioner} and a \emph{Solver}. Given an unlabeled video $v$, the Questioner $\pi_Q$ generates a question $q \sim \pi_Q(\cdot \mid v)$, while the Solver $\pi_S$ produces $M$ candidate answers $\{\hat{a}_m\}_{m=1}^M$, where $\hat{a}_m \sim \pi_S(\cdot \mid v, q)$.  A pseudo target answer $a^*$ is obtained via majority voting:

 \begin{equation}                                         
  a^{*} = \arg\max_{a}\sum_{m=1}^{M}\mathds{1}\left[\hat{a}_{m} = a\right].              
  \label{mv}                                               
\end{equation} 

The goal is to jointly optimize $\pi_Q$ and $\pi_S$ through RL process such that the questioner produces informative and learnable questions, while the solver improves its ability to answer them.

\paragraph{Preliminaries.} 
In existing self-evolving frameworks~\cite{huang2025r, yang2025spell, he2025visplay}, both policies are optimized with rewards derived from solver responses. The solver's confidence on $(v, q)$ is the empirical agreement rate of its $M$ candidates with the majority-voted answer $a^{*}$:                                                              
\begin{equation}                                          
s(v, q) = \frac{1}{M}\sum_{m=1}^{M}\mathds{1}\left[\hat{a}_{m} = a^{*}\right]                  
\label{eq:confidence}                      
\end{equation} 

The questioner is rewarded by a difficulty term:
\begin{equation}
r^{Q}_\mathrm{diff} = \min\big(s(v, q),\, 1 - s(v, q)\big),
\end{equation}
which favors questions of moderate difficulty, along with a diversity penalty $r^{Q}_\mathrm{div}$ that discourages semantically similar questions within a batch, encouraging broader exploration. The overall questioner reward is:
\begin{equation}
r^{Q} = \max\big(0,\, r^{Q}_\mathrm{diff} - r^{Q}_\mathrm{div}\big).
\end{equation}

The solver is trained using a pseudo target derived from Eq. \eqref{mv} and receives a reward composed of an answer correctness term $\mathrm{acc} \in \{0,1\}$ and a format term $\mathrm{fmt} \in \{0,1\}$:
\begin{equation}   
r^{S} = (1 - w)\,\mathrm{acc} + w\,\mathrm{fmt}.
\end{equation}

Both the questioner and solver are optimized using GRPO~\cite{shao2024deepseekmath}, which normalizes rewards within each batch and incorporates a KL regularization term for stable updates. More details and training algorithm can be found in the Appendix \ref{app:detail} and \ref{app:algo}.

\begin{figure}[t]
    \centering
    \includegraphics[width=\linewidth]{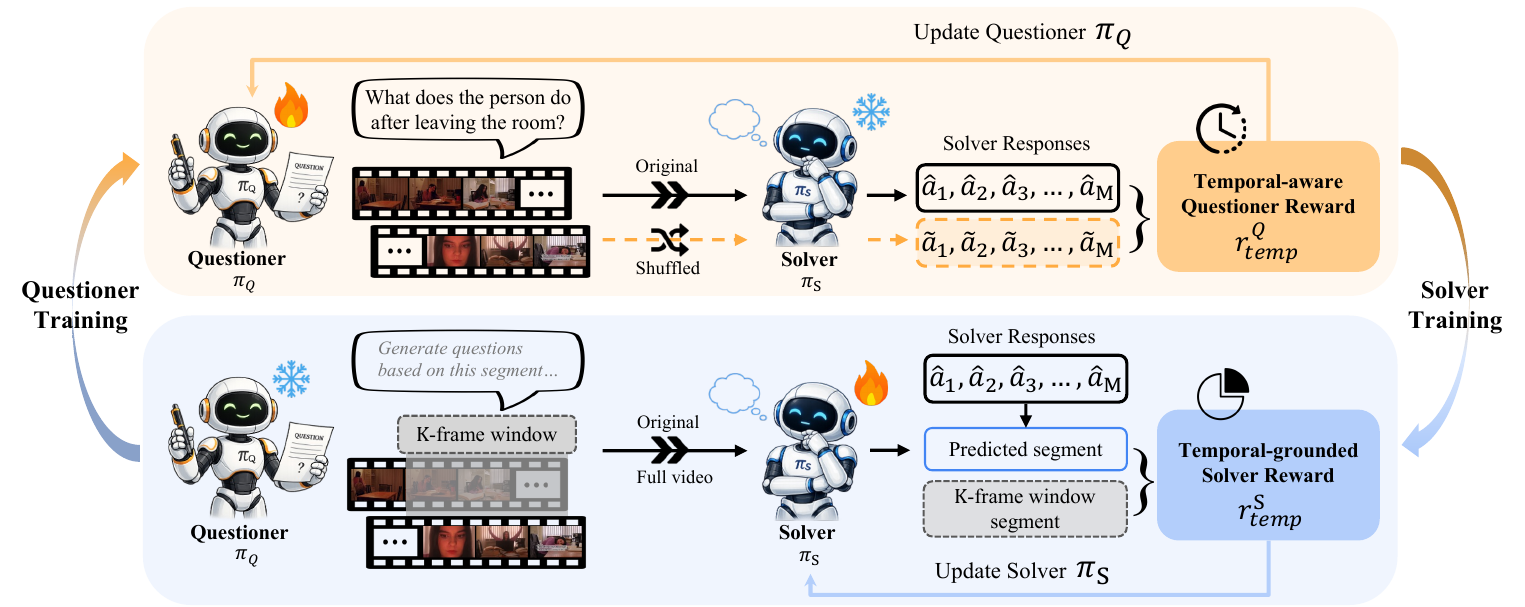}
    \caption{\textbf{Overview of \ours{}.} Questioner $\pi_{Q}$ and Solver $\pi_{S}$ co-evolve through two temporal-centric rewards. \textbf{Questioner Training:} with $\pi_{S}$ frozen, $\pi_{Q}$ generates questions while $\pi_{S}$ responds using original and shuffled frames to derive the temporal-aware Questioner reward $r^{Q}_{\mathrm{temp}}$. \textbf{Solver Training:} with $\pi_{Q}$ frozen, the Questioner generates questions from a sampled $K$-frame window, and $\pi_{S}$ predicts both the answer and temporal segment, which is compared with the sampled window to derive the temporal-grounded Solver reward $r^{S}_{\mathrm{temp}}$. Preliminary rewards are omitted for simplicity.}  
  \label{fig:framework}                                        
\end{figure}
  
\subsection{Temporal-aware Questioner}
The reward signals in existing self-evolving frameworks, introduced in \S\ref{sec:prelim}, are largely modality-agnostic and do not explicitly capture the temporal structure of video. In particular, the preliminary Questioner rewards $r^{Q}_\mathrm{diff}$ and $r^{Q}_\mathrm{div}$ depend only on $(q, s(v, q))$, providing no signal that emphasizes temporal dependencies across frames. As a result, the Questioner is confined to a temporally insensitive regime and tends to behave similarly to image-based question generation, producing questions that can be answered from a single frame. It therefore receives little incentive to explore temporal relationships such as order, causality, or state transitions, ultimately limiting the effectiveness and expressiveness of the learned supervision.

To address this, we introduce a temporal-aware Questioner reward that explicitly leverages the intrinsic supervision provided by the ordering of video frames. Given a question $q$, we evaluate the Solver on both the original video $v$ and a temporally perturbed version $\tilde{v}$ obtained by applying a permutation $\sigma$ to the frame sequence, \ie, $\tilde{v} = \sigma(v)$. Let $s(v, q)$ and $s(\tilde{v}, q)$ denote the corresponding confidence scores defined in Eq.~\eqref{eq:confidence}. A temporally meaningful question should depend on the correct frame order and therefore lead to degraded Solver confidence under temporal perturbation. We define the temporal-aware reward as:
\begin{equation}
r^{Q}_{\mathrm{temp}} = \max\big(0,\, s(v, q) - s(\tilde{v}, q)\big).
\end{equation}
This reward favors questions whose answers become less consistent when temporal order is disrupted, thereby encouraging the Questioner to generate queries that are sensitive to temporal structure. The final Questioner reward is:
\begin{equation}
r^{Q}_{\mathrm{total}} = r^{Q} + \lambda_q\, r^{Q}_{\mathrm{temp}},
\end{equation}
where $\lambda_q$ controls the strength of temporal supervision. By explicitly introducing temporal perturbations during training, the proposed reward promotes the generation of questions that require reasoning over action sequences and event relations, guiding learning toward temporally and causally informative content.

\subsection{Temporal-grounded Solver}
While the temporal-aware Questioner reward improves the temporal sensitivity of generated questions, the Solver itself remains weakly supervised with respect to temporal structure. In the preliminary rewards, Solver is optimized only for answer correctness, without explicit guidance on when relevant evidence occurs in the video. Since the base reward $r^{S}$ is invariant to the temporal location of evidence, a correct answer that ignores temporal locality is rewarded identically to one that accurately localizes supporting evidence. Consequently, Solver receives no incentive to align its predictions with specific temporal segments, limiting its ability to perform temporally grounded reasoning.

To address this, we introduce a temporally grounded supervision signal by exploiting the inherent characteristic of video modality within the self-play pipeline. During question generation used for Solver training, each video is sampled as a contiguous $K$-frame window, which defines a temporal segment $\mathcal{W} = [t_s, t_e]$ in the original video. Since Questioner generates queries conditioned on this window, $\mathcal{W}$ naturally serves as an automatically constructed pseudo ground-truth segment.
In addition to predicting the answer, the Solver is required to output a temporal segment $\hat{\mathcal{W}} = [\hat{t}_s, \hat{t}_e]$ in format \texttt{<segment>$\hat{t}_s$s-$\hat{t}_e$s</segment>}, and is rewarded based on its alignment with $\mathcal{W}$ using Intersection-over-Union (IoU):
\begin{equation}
r^{S}_{\mathrm{temp}} = \mathrm{IoU}(\hat{\mathcal{W}}, \mathcal{W}).
\end{equation}
To ensure meaningful grounding, this reward is applied only when the predicted answer is correct. The final Solver reward is:
\begin{equation}
r^{S}_{\mathrm{total}} = r^{S} + \lambda_s \, r^{S}_{\mathrm{temp}},
\end{equation}
where $\lambda_s$ balances answer correctness and temporal grounding. By aligning predictions with automatically derived temporal segments, Solver is encouraged to learn temporally localized reasoning without requiring human annotations.

\section{Experiments}
\label{sec:exp}

\subsection{Experimental Setup}
\paragraph{Training Data.}
During self-evolving training, we aggregate a diverse collection of open-domain video data. The raw videos are sourced from the training splits of LLaVA-Video-178K~\citep{zhang2024llava}, STAR~\citep{wu2024star}, CLEVRER~\citep{yi2019clevrer}, PerceptionTest~\citep{patraucean2023perception}, NeXT-QA~\citep{xiao2021next}, and Video-Holmes~\citep{videoholmes_2025}, resulting in a total of 5,778 videos. Importantly, we do not use any annotations from these datasets during training, only the raw videos are utilized.

\paragraph{Evaluation Benchmarks.}
We evaluate our model on six video benchmarks spanning both reasoning-focused and general-purpose video understanding tasks. The reasoning benchmarks include Video-Holmes~\citep{videoholmes_2025}, VSIBench~\citep{vsibench_2025}, VideoMMMU~\citep{videommmu_2025}, and MMVU~\citep{mmvu_2025}, which assess the model’s reasoning capabilities in video understanding. The general-purpose benchmarks, TempCompass~\citep{tempcompass_2024} and VideoMME~\citep{videomme_2024}, evaluate performance across a mixture of perception and reasoning tasks. For MMVU, we follow previous works~\cite{feng2025video, park2025deepvideo, wang2026video} to evaluate on its multiple-choice questions. For VideoMME, no subtitles are used during evaluation.

\paragraph{Implementation Details.}
We apply our method to four base models: Qwen2.5-VL-3B-Instruct~\citep{bai2025qwen25vl}, Qwen2.5-VL-7B-Instruct~\citep{bai2025qwen25vl}, Qwen3-VL-4B-Instruct~\citep{bai2025qwen3vl}, and Qwen3-VL-8B-Instruct~\citep{bai2025qwen3vl}. Both the Questioner $\pi_Q$ and Solver $\pi_S$ are initialized from the same base model and trained separately using LoRA fine-tuning. The input is limited to at most 16 frames with a resolution of $128 \times 14 \times 14$. During GRPO~\citep{shao2024deepseekmath} optimization, we use a global batch size of 128 with group size $G{=}8$ and a KL coefficient of $10^{-2}$ for both Questioner and Solver training. For Questioner training, each candidate question is evaluated by sampling $M=10$ answers from the current Solver to compute the reward. The co-evolution process runs for three rounds, alternating between Questioner and Solver, with each phase trained for 20 steps. We set the reward weights to $\lambda_q = 0.1$ for the temporal-aware Questioner reward and $\lambda_s = 0.3$ for the temporal-grounded Solver reward, with a grounding window of $K = 8$ frames. The Questioner and Solver are optimized using AdamW with learning rates of $10^{-6}$ and $2\times10^{-6}$ for 20 steps across three co-evolution rounds. Experiments are conducted with $8 \times$ H100 GPUs. At evaluation time, we adopt the official Qwen2.5-VL decoding settings (\texttt{top\_p} = 0.001, \texttt{temperature} = 0.01). More details can be found in the Appendix~\ref{app:detail}.

\subsection{Main Results}
\begin{table*}[t]                                          
  \centering                                              
  \setlength{\tabcolsep}{3.5pt}             
  \footnotesize                                       
  \caption{\textbf{Performance on video reasoning and general benchmarks across four MLLMs.} For each base model, we compare the untrained \emph{Base Model} against the frozen-questioner (\textcolor{cyan!40}{\faSnowflake} Questioner) baseline, where Solver is trained on questions emitted by an \emph{untrained} Questioner, and three iterations of \ours{} co-evolution. \textbf{Bold} marks the best result for each model. \textcolor{cyan!40}{Blue} superscripts on the Iter-3 row report the absolute improvement over the corresponding base model.}        
  \label{tab:main}                        
  \begin{tabular}{lcccccc}                              
  \toprule                                        
  & \multicolumn{4}{c}{\textbf{Video Reasoning Benchmark}} & \multicolumn{2}{c}{\textbf{Video General Benchmark}} \\   
  \cmidrule(lr){2-5} \cmidrule(lr){6-7}                   
  \textbf{Methods} & \textbf{Video-Holmes} & \textbf{VSI-Bench} & \textbf{VideoMMMU} & \textbf{MMVU} & \textbf{TempCompass} & \textbf{VideoMME} \\
  
  \midrule                                                   
  \multicolumn{7}{l}{\emph{Qwen2.5-VL-3B-Instruct}} \\     \rowcolor{gray!20}               
  \quad Base Model (w/o training) & 26.8 & 25.1 & 30.6 & 48.5 & 28.9 & 48.0 \\
  \quad \ours ({\color{cyan!40}{\faSnowflake}} Questioner) & 27.8 & 26.3 & 35.0 & 53.3 & 34.5 & 48.3 \\             
  \quad \ours (Iter 1) & \textbf{28.0} & 29.5 & 34.4 & 51.7 & 33.2 & 47.0 \\                                     
  \quad \ours (Iter 2) & 26.9 & 28.4 & \textbf{36.9} & 53.1 & 46.3 & \textbf{48.9} \\
  \quad \ours (Iter 3) & 27.2\bgap{0.4} & \textbf{29.5}\bgap{4.4} & 36.7\bgap{6.1} & \textbf{54.6}\bgap{6.1} & \textbf{48.2}\bgap{19.3} & 48.6\bgap{0.6} \\ 
  
  \midrule                                                   
  \multicolumn{7}{l}{\emph{Qwen2.5-VL-7B-Instruct}} \\     \rowcolor{gray!20}                                 
    \quad Base Model (w/o training) & 27.8 & 27.7 & 47.8 & 59.2 & 72.2 & 53.1\\
  \quad \ours ({\color{cyan!40}{\faSnowflake}} Questioner) & 28.4 & 29.3 & 46.6 & 60.5 & 72.5 & 51.5 \\             
  \quad \ours (Iter 1) & 28.6 & 28.5 & 48.9 & 60.6 & 72.1 & 51.4 \\                           
  \quad \ours (Iter 2) & \textbf{29.9} & 30.9 & 48.3 & 60.5 & 73.1 & \textbf{53.8} \\
  \quad \ours (Iter 3) & 29.3\bgap{1.5} & \textbf{31.7}\bgap{4.0} & \textbf{50.0}\bgap{2.2} & \textbf{62.4}\bgap{3.2} & \textbf{73.2}\bgap{1.0} & 53.6\bgap{0.5} \\ 
  
  \midrule                                        
  \multicolumn{7}{l}{\emph{Qwen3-VL-4B-Instruct}} \\    
  \rowcolor{gray!20} 
   \quad Base Model (w/o training) & 29.9 & 40.1 & 46.7 & 60.3 & 70.9 & 49.6 \\
  \quad \ours ({\color{cyan!40}{\faSnowflake}} Questioner) & 30.7 & 40.6 & 45.1 & 61.1 & 70.6 & 49.6 \\             
  \quad \ours (Iter 1) & 30.4 & 41.4 & 47.3 & 61.0 & 70.8 & 49.7 \\      \quad \ours (Iter 2) & 29.1 & \textbf{43.1} & 47.3 & 62.4 & 70.9 & 49.7 \\
  \quad \ours (Iter 3) & \textbf{31.1}\bgap{1.2} & 42.8\bgap{2.7} & \textbf{47.6}\bgap{0.9} & \textbf{62.6}\bgap{2.3} & \textbf{71.7}\bgap{0.8} & \textbf{51.4}\bgap{1.8} \\ 
  
  \midrule                                        
  \multicolumn{7}{l}{\emph{Qwen3-VL-8B-Instruct}} \\ 
  \rowcolor{gray!20}
  \quad Base Model (w/o training) & 36.2 & 37.4 & 46.8 & 64.8 & 73.6 & 50.3 \\
  \quad \ours ({\color{cyan!40}{\faSnowflake}} Questioner) & 36.3 & 35.3 & 47.8 & 63.7 & 73.4 & 50.4 \\             
  \quad \ours (Iter 1) & 36.0 & 38.0 & 49.0 & 63.8 & 74.3 & 50.3 \\                                     
  \quad \ours (Iter 2) & 35.2 & 38.3 & 48.1 & \textbf{67.0} & 73.7 & \textbf{51.4} \\
  \quad \ours (Iter 3) & \textbf{36.6}\bgap{0.4} & \textbf{39.8}\bgap{2.4} & \textbf{49.1}\bgap{2.3} & 66.2\bgap{1.4} & \textbf{74.3}\bgap{0.7} & 51.2\bgap{0.9} \\                          
  \bottomrule                                     
  \end{tabular}                                   
  \end{table*}    
Table~\ref{tab:main} presents the results on six video reasoning and general benchmarks using four MLLMs, including Qwen2.5-VL-Instruct 3B/7B and Qwen3-VL-Instruct 4B/8B. Consistent performance improvements are observed across different base models throughout the training iterations of our proposed self-evolving framework \ours. The gains across all benchmarks validate the effectiveness of our approach despite the absence of human-curated questions or annotations. Through a fully autonomous learning process, the model progressively improves its video understanding and reasoning capabilities by independently generating and solving training questions. For the Frozen Questioner setting, where the solver is trained for three iterations (60 steps) using questions generated by an untrained base model, the achievable improvements are naturally constrained by the quality of the initial policy. Among all models, Qwen2.5-VL-3B exhibits the most significant gains on temporally challenging benchmarks, particularly TempCompass (+19.3\%), VideoMMMU (+6.1\%), and MMVU (+6.1\%), highlighting the effectiveness of our framework even when starting from a relatively weaker initialization. Across all model scales, larger improvements are consistently observed on video reasoning benchmarks that require temporal understanding, aligning with our design motivation of explicitly optimizing temporally grounded reward signals.

\subsection{Comparison with Supervised Baselines and \textsc{Vanilla} Self-evolving Baseline}
\begin{table}[t]
    \centering
    \setlength{\tabcolsep}{2.5pt}
    \footnotesize
    \caption{\textbf{Comparison with supervised baselines and \textsc{Vanilla} self-evolving baseline.} We evaluate \ours on Qwen2.5-VL-7B against three reference approaches: supervised fine-tuning (\textsc{SFT}), supervised RL with GRPO (\textsc{Vanilla GRPO}), and the label-free \textsc{Vanilla} self-evolving framework without temporal-inherent rewards. $^{\dagger}$ denotes methods trained on human-curated datasets with both questions and annotations. In contrast, both \textsc{Vanilla} self-evolving and \ours are trained directly from raw videos without human supervision. We also initialize \ours from the SFT checkpoint, with results reported in the last row. \textbf{Bold} and \underline{underline} indicate the best and second-best results.}
    \label{tab:baselines}
     \begin{tabular}{lccccccc}                              
  \toprule                                        
  & \multicolumn{4}{c}{\textbf{Video Reasoning Benchmark}} & \multicolumn{2}{c}{\textbf{Video General Benchmark}} & \\   
  \cmidrule(lr){2-5} \cmidrule(lr){6-7}                   
  \textbf{Methods} & \textbf{Video-Holmes} & \textbf{VSI-Bench} & \textbf{VideoMMMU} & \textbf{MMVU} & \textbf{TempCompass} & \textbf{VideoMME} & \textbf{Avg.} \\
  
  \midrule 
  \rowcolor{gray!20}
  Base Model (w/o training) & 27.8 & 27.7 & 47.8 & 59.2 & 72.2 & 53.1 & 47.9\\
  SFT$^{\dagger}$ & 31.1 & 31.8 & 47.4 & 61.3 & 69.2 & 52.8 & 49.0\\ 
  \textsc{SFT$^{\dagger}$ + Vanilla GRPO}$^{\dagger}$ & \textbf{38.8} & \underline{32.7} & 48.3 & 62.1 & 71.3 & \textbf{54.5} & \textbf{51.3} \\
  \textsc{Vanilla} Self-evolving & 28.9 & 28.9 & \underline{49.4} & 59.2 & \underline{72.9} & 52.5 & 48.6\\
  \rowcolor{cyan!10}
  \ours (Iter 3) & 29.3 & 31.7 & \textbf{50.0} & \underline{62.4} & \textbf{73.2} & 53.6 & 50.0 \\
  \rowcolor{cyan!10}
  SFT$^{\dagger}$ + \ours (Iter 3) & \underline{33.3} & \textbf{35.0} & 48.6 & \textbf{63.5} & 72.6 & \underline{54.0} & \underline{51.2} \\                                         
  \bottomrule           
  \end{tabular} 
  \vspace{-1em}
\end{table}

Table~\ref{tab:baselines} compares \ours based on Qwen2.5-VL-7B against supervised fine-tuning (SFT), supervised RL with GRPO (\textsc{vanilla} GRPO), and the \textsc{vanilla} self-evolving baseline. As shown in the results, \ours surpasses SFT by an additional average gain of +1.0\%, where SFT is trained for one epoch on the Video-R1-CoT-165k dataset~\cite{feng2025video}. Applying an additional 1k steps of \textsc{vanilla} GRPO further improves performance over SFT by 2.3\%, while our method achieves a 2.1\% improvement over the base model without relying on any human annotations. Compared with the \textsc{vanilla} self-evolving baseline, our approach increases the average performance from 48.6\% to 50.0\%, achieving the best results across all evaluated datasets. Notably, among all baselines, including supervised methods, \ours achieves the best performance on VideoMMMU, MMVU, and TempCompass, demonstrating the effectiveness of the self-evolving framework for video understanding through the integration of temporal-centric rewards for both the Questioner and Solver. We further initialize \ours with Video-R1 SFT checkpoint \cite{feng2025video}. The resulting model achieves performance competitive with the supervised \textsc{SFT + vanilla GRPO} baseline while requiring significantly fewer RL training steps. This highlights the potential of our framework to further enhance pretrained Video-LLMs through self-evolving temporal supervision without introducing additional human annotations.

\subsection{Ablation Studies}
\begin{table}[t]
    \centering
    \setlength{\tabcolsep}{5pt}
    \footnotesize
    \caption{\textbf{Reward ablation on proposed temporal Questioner and Solver rewards.} We conduct ablation based on Qwen2.5-VL-7B at Iter 3 on the proposed temporal-aware Questioner reward $r^{Q}_{\mathrm{temp}}$ and temporal-grounded Solver reward $r^{S}_{\mathrm{temp}}$. The first row represents \textsc{vanilla} Self-evolving with only preliminary rewards present in \S\ref{sec:prelim}.}
    \label{tab:ablation}
    \begin{tabular}{ccccccccc}                             
    \toprule 
   \multicolumn{2}{c}{\textbf{Reward}}& \multicolumn{4}{c}{\textbf{Video Reasoning Benchmark}} & \multicolumn{2}{c}{\textbf{Video General Benchmark}} & \\ 
   
  \cmidrule(r){1-2} \cmidrule(r){3-6} \cmidrule(r){7-8}
  \textbf{$r^{Q}_{temp}$} & \textbf{$r^{S}_{temp}$} & \textbf{Video-Holmes} & \textbf{VSI-Bench} & \textbf{VideoMMMU} & \textbf{MMVU} & \textbf{TempCompass} & \textbf{VideoMME} & \textbf{Avg.} \\
  
  \midrule
  \xmarkg& \xmarkg  & 28.9 & 28.9 & 49.4 & 59.2 & 72.9 & 52.5 & 48.6 \\
  \cmark& \xmarkg  & 29.7 & 29.6 & 49.1 & 62.1 & 72.8	& 52.7 & 49.3 \\ 
  \xmarkg& \cmark & \textbf{30.1} & 30.7 & 49.3 & 61.0 & 71.8 & 52.5 & 49.2 \\ 
  \rowcolor{cyan!10}
  \cmark& \cmark  & 29.3	& \textbf{31.7}	& \textbf{50.0} & \textbf{62.4} & \textbf{73.2} & \textbf{53.6} & \textbf{50.0} \\                                
  \bottomrule           
  \end{tabular} 
  \vspace{-1em}
\end{table}

\paragraph{Reward Ablation.} Table~\ref{tab:ablation} isolates the contributions of the two temporal-inherent rewards by incrementally adding each component to the \textsc{Vanilla} Self-evolving baseline. Using only the temporal-aware Questioner reward $r^{Q}_{\mathrm{temp}}$ improves the average performance across the six benchmarks by +0.7\%, while using only the temporal-grounded Solver reward $r^{S}_{\mathrm{temp}}$ increases the average to 49.2\%, corresponding to a +0.6\% gain. Combining both rewards produces the full model, achieving 50.0\% average performance, which surpasses \textsc{Vanilla} self-evolving by +1.4\%. The complementary improvements indicate that the gains from the two rewards are additive rather than redundant, consistent with their distinct optimization roles on different agents and temporal aspects. Specifically, the temporal-aware Questioner reward guides the questioner through frame-order sensitivity, whereas the temporal-grounded Solver reward enhances the solver via temporally consistent segment localization.

\begin{figure}[t]
    \centering
    \begin{minipage}{0.48\textwidth}
        \centering
        \includegraphics[width=\linewidth]{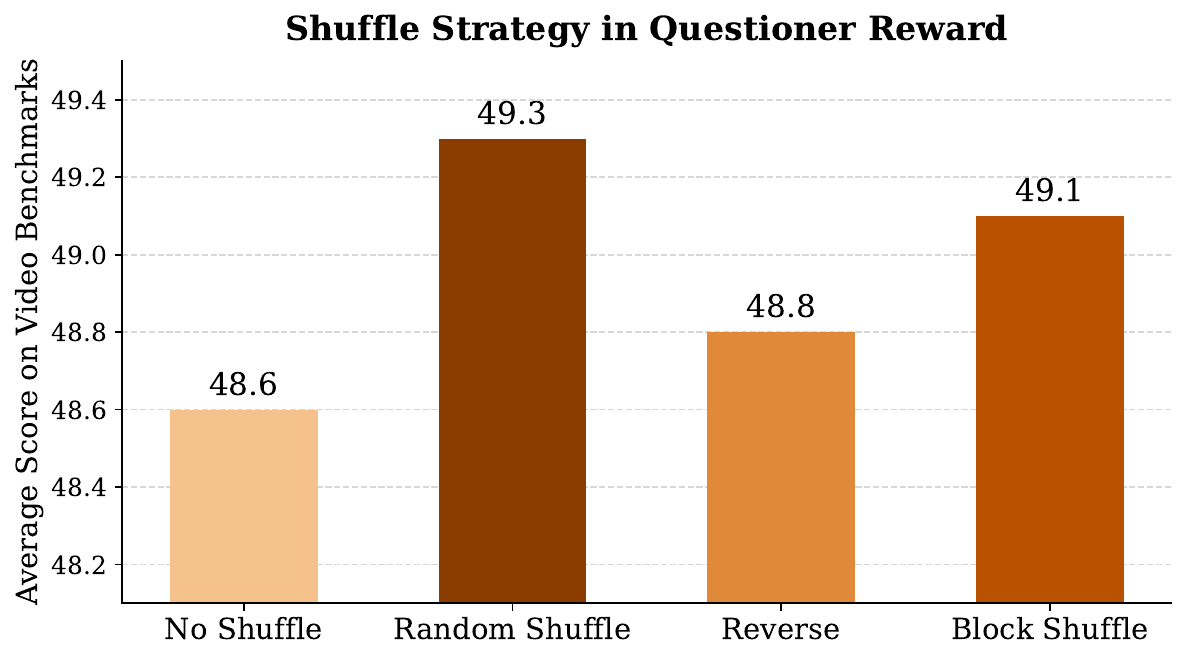}
        \caption{Effect of different shuffle strategies in Questioner reward on performance. We use random shuffle by default.}
        \label{fig:shuffle}
    \end{minipage}
    \hfill
    \begin{minipage}{0.48\textwidth}
        \centering
        \includegraphics[width=\linewidth]{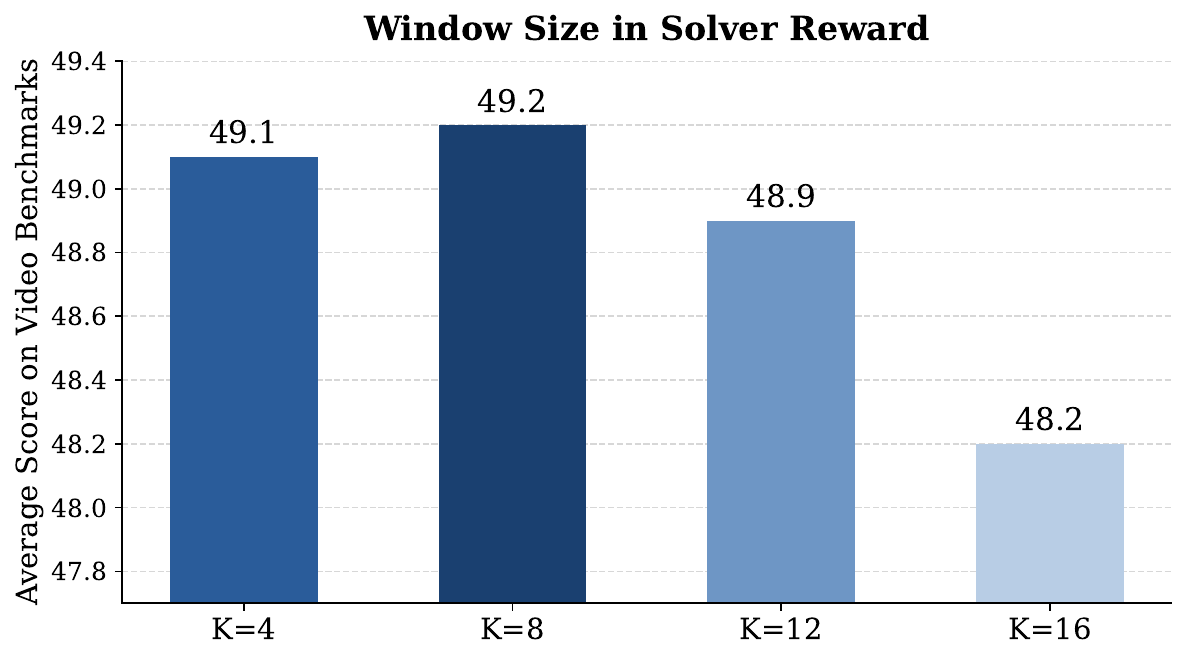}
        \caption{Effect of different window sizes in Solver reward on performance. We use $K=8$ by default.}
        \label{fig:winsize}
    \end{minipage}
    \vspace{-0.5em}
\end{figure}

\paragraph{Shuffle Strategy in Questioner Reward.}
Figure~\ref{fig:shuffle} illustrates the impact of different frame-shuffling strategies on model performance. We compare a no-shuffle baseline with three temporal perturbation strategies: random shuffle (default), reverse-order frames, and block-wise random shuffle ($\text{block}=4$). All shuffling strategies outperform the no-shuffle baseline (48.6\%), indicating that temporally degraded contrastive signals are effective for encouraging temporal sensitivity during training. Among the variants, simply reversing the frame order yields only a modest improvement of +0.2\%, as it perturbs only the temporal direction while preserving much of the original temporal structure. Both block shuffle and fully random shuffle achieve stronger and comparable gains, although block shuffle performs slightly worse because it retains short-range temporal coherence while mainly disrupting long-range dependencies. In contrast, random shuffle removes temporal structure more thoroughly, providing a stronger temporal contrastive signal. We use random shuffle in all experiments by default.

\paragraph{Window Size in Solver Reward.}
Figure~\ref{fig:winsize} compares the effect of different window sizes $K$ used for question generation and temporal-grounded solver supervision. We sweep $K \in \{4, 8, 12, 16\}$ on 16-frame videos. For moderate window sizes ($K=\{4,8,12\}$), performance remains stable, achieving 48.9–49.2 on average with only 0.3-point variation, indicating that the method is not highly sensitive to the exact window size within an informative regime. Performance drops at $K=16$, where the sampled segment spans the full video and the IoU-based grounding bonus becomes degenerate (48.2\%). This result confirms that the gain mainly comes from informative temporal segment supervision rather than merely introducing a \texttt{<segment>} output. Overall, the results suggest that any reasonable window size below the full video length can provide effective grounding supervision. We adopt $K=8$ as the default setting in all experiments.

\subsection{Qualitative Analysis of Generated Questions}
\begin{figure}[t]
    \centering
    \includegraphics[width=\linewidth]{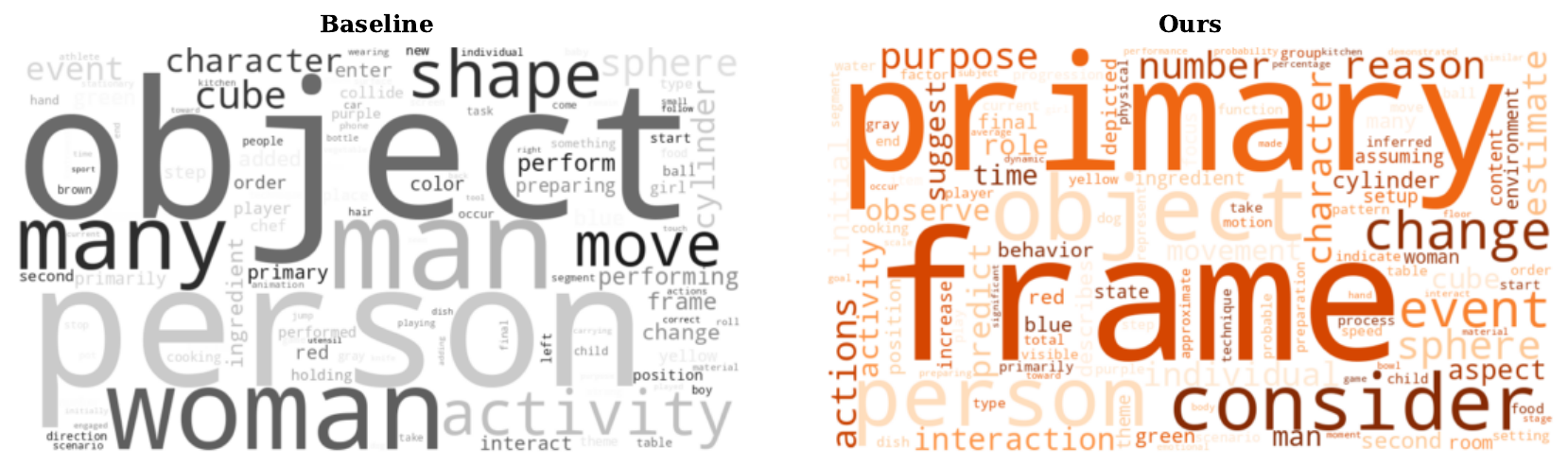}
    \caption{{\textbf{Qualitative analysis of generated questions.} The word cloud visualization shows that \ours{} generates questions with substantially richer temporal and reasoning-oriented vocabulary compared to the \textsc{vanilla} Self-evolving baseline.}}  
  \label{fig:wordcloud}
  \vspace{-1em}
\end{figure}

Figure~\ref{fig:wordcloud} visualizes the word distributions of generated questions between the \textsc{vanilla} Self-evolving baseline and our temporal-aware Questioner, revealing a sharp lexical contrast. Compared with the baseline, which is dominated by generic and appearance-oriented terms such as \emph{person}, \emph{object}, and \emph{shape}, our model produces substantially richer temporal and reasoning-related vocabulary. In particular, words such as \emph{primary}, \emph{frame}, \emph{consider}, \emph{change}, \emph{reason} and \emph{interaction} appear much more frequently, indicating that the generated questions increasingly focus on temporal transitions, causal relations, and dynamic scene understanding. The increased diversity of temporal and relational concepts suggests that the proposed temporal-aware Questioner reward successfully encourages the Questioner to generate more temporally dependent and reasoning-intensive queries, rather than relying on general and static visual attributes alone.

\subsection{Iteration Scaling}
\begin{wrapfigure}[14]{r}{0.48\textwidth}
    \vspace{-1.2em}
    \includegraphics[width=0.98\linewidth]
    {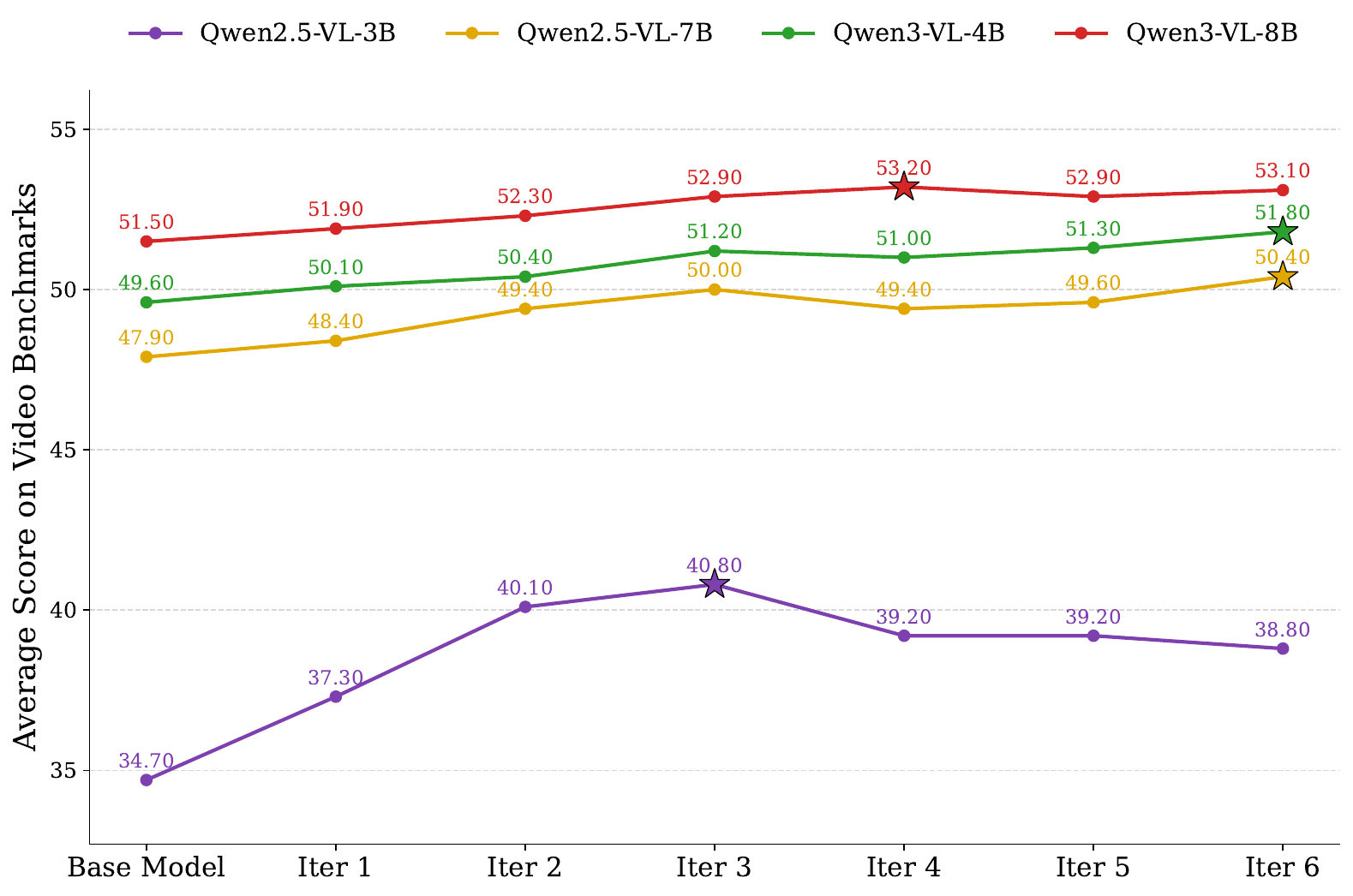}
    \vspace{-0.5em}
    \caption{\textbf{Iteration scaling.} Performance under extended training iterations and model scales.}
    \label{fig:itercurve}
\end{wrapfigure}
Figure~\ref{fig:itercurve} extends the co-evolution process beyond the default three iterations. Three of the four backbones (Qwen2.5-VL-7B, Qwen3-VL-4B, and Qwen3-VL-8B) exhibit a generally stable increasing trend, where performance improves progressively with additional iterations despite minor fluctuations at certain stages. Overall, the curves show a clear upward trajectory, indicating that temporal-centric self-evolution continues to provide meaningful gains during iterative training. In contrast, Qwen2.5-VL-3B behaves less consistently, reaching its peak performance at iteration 3 before showing slight degradation in later iterations. We attribute this instability to the limited capacity of Qwen2.5-VL-3B model, which may hinder its ability to sustain long-term co-evolution optimization. Nevertheless, the results suggest that the default three-iteration setting used in our experiments demonstrates balanced performance-cost tradeoff while maintaining stable training across different model scales.

\subsection{Question Evolution}
\begin{wraptable}[10]{r}{0.48\textwidth}
    \vspace{-1.2em}
    \caption{\textbf{Question evolution.} Solver performance evaluated on questions generated by Questioner across different iterations.}
    \setlength{\tabcolsep}{10pt}                                     
  \footnotesize
  \begin{tabular}{lccc}                       
  \toprule                                                   
  & $\mathcal{D}_\text{Iter 1}$ & $\mathcal{D}_\text{Iter 2}$ & $\mathcal{D}_\text{Iter 3}$ \\
  \midrule                                                       
  Base Model    & 33.0 & 19.0 & 13.5 \\                       
  Solver@Iter 1 & 35.5 & 20.5 & 17.0 \\   
  Solver@Iter 2 & 36.5 & 23.5 & 18.5 \\                      
  Solver@Iter 3 & 38.0 & 24.5 & 22.0 \\   
  \bottomrule                                                    
  \end{tabular}

    \label{tab:question}
\end{wraptable}
Table~\ref{tab:question} investigates the evolution of Questions generated by Questioner across iterations. We sample 300 questions generated at each iteration, denoted as $\mathcal{D}_\text{Iter}$, and evaluate Solver accuracy using GPT-5.4-derived responses as \emph{oracle} answers. Reading across rows, the accuracy decreases substantially (e.g., from 33.0\% to 13.5\% for the base model), indicating that Questioner progressively constructs a more challenging curriculum over iterations. Reading down columns, each successive Solver consistently outperforms its predecessor, demonstrating that Questioner and Solver co-evolve jointly, with Solver capability improving alongside the increasing difficulty of generated questions.

\section{Conclusion}
\label{sec:conclusion}
In this work, we present \ours, a temporal-centric self-evolving framework for video understanding and reasoning. Driven by the intrinsic temporal dynamics of video, we design a temporal-aware Questioner and a temporal-grounded Solver that co-evolve through iterative training, yielding consistent improvements across diverse benchmarks and model scales. Our work points to a new paradigm for Video-LLMs, enabling scalable and fully autonomous learning.

\pagebreak


{\small
\bibliographystyle{unsrt}
\bibliography{sections/11_references}
}

\newpage
\appendix \section*{Appendix}
\section{Implementation Details}
\label{app:detail}

\subsection{Reward Implementation}
\label{app:rewards}

The full Questioner and Solver rewards are defined in \S\ref{sec:method}. Here we give the implementation-level details that were left implicit in the main paper.

\paragraph{Format reward.}
For both the Questioner and Solver, we apply a format gate before assigning any reward. Questioner outputs that do not contain a parseable \texttt{<type>}/\texttt{<question>}/\texttt{<answer>} triple receive zero reward. Similarly, Solver outputs without a \texttt{\textbackslash boxed{}} answer receive zero accuracy reward, and outputs lacking a parseable \texttt{<segment>$t_s$--$t_e$</segment>} yield zero IoU bonus, preventing malformed predictions from corrupting the gradient. Concretely, the indicator $\mathds{1}^{Q}_{\mathrm{fmt}}\in\{0,1\}$ in Algorithm~\ref{alg:chronos} denotes the Questioner format gate, while $r_{\mathrm{fmt}}\in\{0,1\}$ in the Solver branch serves as the Solver format gate. The format weight $w$ in Eq.~\eqref{eq:rs} further balances correctness against format compliance.

\paragraph{Confidence and the difficulty term.}
The Solver confidence $s(v,q)$ is computed exactly as in Eq.~\eqref{eq:confidence} of the main paper, with $M{=}10$ Solver samples per question. We re-state the difficulty term for clarity:
\begin{equation}
r^{Q}_{\mathrm{diff}} = \min\big(s(v, q),\, 1 - s(v, q)\big).
\end{equation}

\paragraph{Diversity penalty.}
For each video $v$, the Questioner produces a group of $G$ rollouts $\{q_g\}_{g=1}^{G}$ (with $G{=}8$ following the GRPO group size). We compute pairwise BLEU similarities among the group and run agglomerative clustering with average linkage at a fixed BLEU threshold of $\tau_{\mathrm{BLEU}}{=}0.5$ to obtain clusters $\{\mathcal{C}_k\}$. The diversity penalty for a question $q$ in cluster $\mathcal{C}_k$ is
\begin{equation}
r^{Q}_{\mathrm{div}} = \lambda_d \,\cdot\, \frac{|\mathcal{C}_k|}{G},\qquad \lambda_d=1,
\end{equation}
so that questions inside large clusters (i.e.\ the Questioner is repeating itself across the group) are penalized more heavily. The combined preliminary Questioner reward used in the main paper (\S\ref{sec:prelim}) is then $r^{Q} = \max\big(0,\, r^{Q}_{\mathrm{diff}} - r^{Q}_{\mathrm{div}}\big)$.

\paragraph{Final Questioner and Solver rewards.}
Combining the preliminary terms with the temporal-aware Questioner reward $r^{Q}_{\mathrm{temp}}$ and the temporal-grounded Solver reward $r^{S}_{\mathrm{temp}}$ from \S\ref{sec:method}, the rewards used in GRPO are:
\begin{align}
r^{Q}_{\mathrm{total}} &= \mathds{1}^{Q}_{\mathrm{fmt}} \cdot \big(\,r^{Q} + \lambda_q\, r^{Q}_{\mathrm{temp}}\,\big), \label{eq:rq}\\
r^{S}_{\mathrm{total}} &= (1{-}w)\,\mathrm{acc} + w\,\mathrm{fmt} + \lambda_s\, r^{S}_{\mathrm{temp}}, \label{eq:rs}
\end{align}
where a gating $\cdot \mathrm{acc}$ is applied to $ r^{S}_{\mathrm{temp}}$ to ensure that grounding is rewarded only when the predicted answer matches the pseudo-target. We use $\lambda_q{=}0.1$ and $\lambda_s{=}0.3$ throughout (\S\ref{sec:exp}); $w{=}0.1$ keeps format a small fraction of the Solver reward.

\subsection{Prompt Templates}
\label{app:prompts}
\paragraph{Questioner prompt.}
The Questioner is given a prompt that constrains the output schema and the admissible question types. We require a single \texttt{<type>}/\texttt{<question>}/\texttt{<answer>} triple per rollout, where the type is exactly one of $\{$\textit{multiple choice}, \textit{numerical}, \textit{regression}$\}$. The strict format makes the rollouts trivially parseable for the reward function and prevents the model from emitting auxiliary commentary that would dilute the supervision signal. We include one in-context example so the format constraint is grounded for the policy. The full prompt is reproduced verbatim below.

\begin{tcolorbox}[
    colback=gray!5,     
    colframe=gray!30, 
    title=\textbf{Questioner Prompt},
    breakable,
    fonttitle=\bfseries,
    coltitle=black,
    left=2mm,
    right=2mm,
    top=1mm,
    bottom=1mm
]
\small
You are an intelligent Question Generator. Your task is to create a question based on the given video.\\[0.3em]
Requirements (must follow exactly):
\begin{enumerate}\itemsep0.05em
\item Watch the video carefully and understand all details across the frames.
\item Generate exactly one question that is directly related to the video content.
\item Choose the question type from only one of: multiple choice (Yes/No or four options A/B/C/D, one correct), numerical (a specific numeric answer), or regression (a continuous value such as a measurement, quantity, or coordinate).
\item The question must require analysis or reasoning, not just description.
\item Provide the correct answer. Include units if applicable.
\item Output strictly in the three-block format below, with nothing else.
\end{enumerate}
Output format:\\[0.2em]
<type>X</type>\\
<question>Y</question>\\
<answer>Z</answer>\\[0.3em]
where X $\in$ \{multiple choice, numerical, regression\}.
\end{tcolorbox}

\paragraph{Solver prompt.}
The Solver receives a thin chain-of-thought wrapper around the Questioner-generated query, with an additional sentence that instructs the Solver to emit the relevant time segment in a parseable form for the temporal-grounded Solver reward.

\begin{tcolorbox}[
    colback=gray!5,     
    colframe=gray!30, 
    title=\textbf{Solver Prompt},
    breakable,
    fonttitle=\bfseries,
    coltitle=black,
    left=2mm,
    right=2mm,
    top=1mm,
    bottom=1mm
]
\small
\{question\} Please reason step by step based on the question and video, and put your final answer within $\backslash$boxed\{\}. Additionally, predict the video time segment (in seconds) that is most relevant to answering this question and output it as <segment>Xs--Ys</segment> (e.g.\ <segment>2.5s--5.0s</segment>).
\end{tcolorbox}

The IoU bonus $\lambda_s\cdot u\cdot\mathrm{acc}$ is computed only when a parseable \texttt{<segment>} tag is present and the answer is correct, and is masked out otherwise (Eq.~\eqref{eq:rs}).

\subsection{Hyperparameters}
\label{app:hparams}

Table~\ref{tab:hparams} lists all hyperparameters used in the main experiments. We do not tune any hyperparameter on the evaluation benchmarks.
\begin{table}[h]
    \centering
    \setlength{\tabcolsep}{5pt}
    \footnotesize
    \caption{\textbf{Training hyperparameters used in the main experiments.}}
    \label{tab:hparams}
    \begin{tabular}{@{}l l l@{}}
        \toprule
        Component & Hyperparameter & Value \\
        \midrule
        Base models       & 4 backbones & Qwen2.5-VL-3B/7B-Instruct, Qwen3-VL-4B/8B-Instruct \\
        Parameter-efficient FT & LoRA              & rank $32$, $\alpha{=}64$, dropout $0.05$, all linear projections \\
        Training GPUs     & 4$\times$H100, FSDP   & one worker per GPU \\
        Rollout GPUs      & 4$\times$H100, vLLM & tensor parallel $1$, one server per GPU \\
        \midrule
        GRPO group size $G$         & Questioner / Solver  & $8$ / $8$ \\
        Solver samples per question & $M$                  & $10$ \\
        Co-evolution iterations     & $T$                  & $3$ (default; ablated up to $8$) \\
        Steps per phase             & Questioner / Solver  & $20$ / $20$ \\
        \midrule
        Frames per video            & $T_{f}$              & $16$ \\
        Visual tokens per frame     &                      & $128$ ($14{\times}14$ patches) \\
        Questioner window length    & $K$                  & $8$ contiguous frames (default) \\
        \midrule
        Temporal-aware Questioner weight  & $\lambda_q$    & $0.1$ \\
        Temporal-grounded Solver weight   & $\lambda_s$    & $0.3$ \\
        Diversity penalty weight          & $\lambda_d$    & $1.0$ \\
        Solver format weight              & $w$            & $0.1$ \\
        BLEU clustering threshold         & $\tau_{\mathrm{BLEU}}$ & $0.5$ \\
        Score-band filter                 & $[s_{\min}, s_{\max}]$ & $[0.3, 0.8]$ \\
        \midrule
        Optimizer         & AdamW               & $\beta_{1}{=}0.9$, $\beta_{2}{=}0.95$ \\
        Learning rate     &     Questioner / Solver                & $1{\times}10^{-6}$ / $2{\times}10^{-6}$ \\
        Global batch size &                     & $128$ \\
        KL coefficient    & $\beta$ (GRPO)      & $10^{-2}$ \\
        Decoding (eval)   & official Qwen2.5-VL & \texttt{top\_p}$=$$0.001$, \texttt{temperature}$=$$0.01$ \\
        \bottomrule
    \end{tabular}
\end{table}

\section{Algorithm and Pipeline}
\label{app:algo}

Algorithm~\ref{alg:chronos} summarizes the complete co-evolution loop of \ours{}, which consists of three key phases: Phase 1: Questioner Training, Phase 2: Solver Dataset Construction, and Phase 3: Solver Training. In Phase 1, the Questioner is optimized to generate temporally dependent queries under the proposed reward signals. In Phase 2, the updated Questioner generates one question per video conditioned on a uniformly sampled $K$-frame contiguous window $[t_s, t_e]$ across the full training set. For each question, the frozen Solver produces $M{=}10$ candidate answers, from which we derive a majority-vote pseudo-answer $a^{*}$ and a confidence score $s(v,q)$. We retain only samples with confidence in $s\in[0.3, 0.8]$ to form the Solver training set $\mathcal{D}_{t}$, where the lower bound filters out questions that are too difficult for the current Solver, and the upper bound removes those that are already trivial, maintaining a balanced training curriculum. In Phase 3, the Solver is trained to answer the generated questions and predict the corresponding temporal segments. The window timestamps $[t_s, t_e]$ travel with the example into Solver training and serve as the pseudo-ground-truth segment $\mathcal{W}$ in the temporal-grounded reward. These three phases are executed iteratively, enabling the Questioner and Solver to co-evolve and progressively improve the model’s temporal reasoning capability without relying on human supervision.

\begin{algorithm}[t]                                                                                                                                                                                          
    \caption{\ours{}: Temporal-Centric Self-Evolution for Video-LLMs.}                                                                                                                                          
    \label{alg:chronos}                                                                                                                                                                                         
    \begin{algorithmic}[1]                                                                                                                                                                                      
    \REQUIRE Initial models $\pi_{Q}, \pi_{S}$ (initialized from a single base policy);                                 
             unlabeled video set $\mathcal{V}$;                
             group size $G$;                                    
             Solver samples $M$;                                       
             window length $K$;                                 
             score band $[s_{\min}, s_{\max}] = [0.3, 0.8]$;    
             reward weights $\lambda_{q}, \lambda_{s}, w$.         
    \ENSURE  Evolved policies $\pi_{Q}$ and $\pi_{S}$.              
    \FOR{\textit{each self-play iteration}}                                                  
        \STATE \texttt{// --- Phase 1: Questioner Training ---} 
        \FOR{\textit{each Questioner training step}}            
            \STATE Sample a video $v \in \mathcal{V}$ and a question group $\{q_{i}\}_{i=1}^{G} \sim \pi_{Q}(\cdot \mid v)$;                                   
            \FOR{\textit{each question $q_{i}$}}                
                \IF{\textit{Format\_Check($q_{i}$) is invalid (e.g., missing $<$type$>$/$<$question$>$/$<$answer$>$ tags)}}         
                    \STATE $r_{i} \leftarrow 0$;                
                \ELSE                                                       
                    \STATE Sample $M$ answers $\{\hat{a}_{m}\}_{m=1}^{M} \sim \pi_{S}(\cdot \mid v, q_{i})$;                                                                                                    
                    \STATE Get pseudo-label $a^{*} \leftarrow \mathrm{MajorityVote}(\{\hat{a}_{m}\})$;                                                                                                      
                    \STATE Compute Solver confidence $s(v, q_{i}) \leftarrow \tfrac{1}{M}\sum_{m=1}^{M}\mathds{1}\{\hat{a}_{m} = a^{*}\}$;                                                                  
                    \STATE \textcolor{black}{Sample shuffled-frame answers $\{\tilde{a}_{m}\}_{m=1}^{M} \sim \pi_{S}(\cdot \mid \sigma(v), q_{i})$;}                                                              
                    \STATE \textcolor{black}{Compute shuffled confidence $s(\sigma(v), q_{i}) \leftarrow \tfrac{1}{M}\sum_{m=1}^{M}\mathds{1}\{\tilde{a}_{m} = a^{*}\}$;}                                     
                    \STATE Compute difficulty $r^{Q}_{\mathrm{diff}} \leftarrow \min\bigl(s(v, q_{i}),\, 1 - s(v, q_{i})\bigr)$;                                                                                
                    \STATE Compute diversity penalty $r^{Q}_{\mathrm{div}}(q_{i})$ via intra-group BLEU clustering;                                                                                             
                    \STATE \textcolor{black}{Compute temporal-aware reward $r^{Q}_{\mathrm{temp}} \leftarrow \max\bigl(0,\, s(v, q_{i}) - s(\sigma(v), q_{i})\bigr)$;}                                            
                    \STATE Final reward: $r_{i} \leftarrow \max\bigl(0,\, r^{Q}_{\mathrm{diff}} - r^{Q}_{\mathrm{div}}\bigr) + \textcolor{black}{\lambda_{q}\, r^{Q}_{\mathrm{temp}}}$;                       
                \ENDIF                                                                                                                                                                                          
            \ENDFOR                                                                                                                                                                                             
            \STATE Update $\pi_{Q}$ via GRPO using rewards $\{r_{i}\}_{i=1}^{G}$;                                                                      
        \ENDFOR                                                          \STATE \texttt{// --- Phase 2: Solver Dataset Construction ---}                   
        \STATE Initialize curated dataset $\mathcal{D} \leftarrow \emptyset$;                           
        \FOR{\textit{each video $v \in \mathcal{V}$}}                                                   
            \STATE Sample window $[t_{s}, t_{e}]$ of $K$ contiguous frames; let $v^{w}$ denote the windowed clip;                                                                                               
            \STATE Sample one question $q \sim \pi_{Q}(\cdot \mid v^{w})$;                              
                \quad\COMMENT{Questioner sees only $v^{w}$ here}                                        
            \STATE Sample $M$ answers $\{\hat{a}_{m}\}_{m=1}^{M} \sim \pi_{S}(\cdot \mid v, q)$;        
            \STATE Get pseudo-label $a^{*} \leftarrow \mathrm{MajorityVote}(\{\hat{a}_{m}\})$;
            \STATE Compute confidence $s(v, q) \leftarrow \tfrac{1}{M}\sum_{m=1}^{M}\mathds{1}\{\hat{a}_{m} = a^{*}\}$;                                                                                         
            \IF{$s(v, q) \in [s_{\min}, s_{\max}]$}          
                \STATE Add $\bigl(v, [t_{s}, t_{e}], q, a^{*}\bigr)$ to $\mathcal{D}$;                  
            \ENDIF                                           
        \ENDFOR                                     

        \STATE \texttt{// --- Phase 3: Solver Training ---}                        
        \FOR{\textit{each minibatch $(v, [t_{s}, t_{e}], q, a^{*}) \in \mathcal{D}$}}                   
            \STATE Sample $G$ Solver responses $\bigl\{(\hat{a}_{m},\, [\hat{t}^{(m)}_{s}, \hat{t}^{(m)}_{e}])\bigr\}_{m=1}^{G} \sim \pi_{S}(\cdot \mid v, q)$;                                                 
            \STATE Compute correctness $\mathrm{acc}_{m} \leftarrow \mathds{1}\{\hat{a}_{m} = a^{*}\}$ and format reward $\mathrm{fmt}_{m}$;                                                                    
            \STATE \textcolor{black}{Compute IoU $u_{m} \leftarrow \mathrm{IoU}\bigl([\hat{t}^{(m)}_{s}, \hat{t}^{(m)}_{e}],\, [t_{s}, t_{e}]\bigr)$;}       
                \quad\COMMENT{window-as-pseudo-segment}                                         
            \STATE Final reward: $r_{m} \leftarrow (1{-}w)\,\mathrm{acc}_{m} \;+\; w\,\mathrm{fmt}_{m} + \textcolor{black}{\lambda_{s}\, u_{m}\, \mathrm{acc}_{m}}$;
            \STATE Update $\pi_{S}$ via GRPO using rewards $\{r_{m}\}_{m=1}^{G}$;                       
        \ENDFOR                                             
    \ENDFOR                                                  

    \RETURN $\pi_{Q},\, \pi_{S}$.                   
    \end{algorithmic}                               
  \end{algorithm}                                 

\section{Dataset Details}
\label{app:datasets}

\subsection{Training Videos}
Following \S\ref{sec:exp}, we aggregate raw video files from the training splits of LLaVA-Video-178K~\citep{zhang2024llava} (2{,}636), STAR~\citep{wu2024star} (991), CLEVRER~\citep{yi2019clevrer} (920), PerceptionTest~\citep{patraucean2023perception} (640), NeXT-QA~\citep{xiao2021next} (358), and Video-Holmes~\citep{videoholmes_2025} (233), totaling $5{,}778$ unique videos. We use \emph{only the raw video files}; all human-written QA pairs, captions, and temporal-segment annotations bundled with these datasets are discarded before training and never enter the pipeline. The mixture spans synthetic physics (CLEVRER), instructional (LLaVA-Video-178K, NeXT-QA), egocentric (PerceptionTest), and narrative (Video-Holmes, STAR) clips, ensuring that the temporal-inherent rewards see a broad range of visual dynamics rather than a single domain.

\subsection{Evaluation Benchmarks}
\label{app:benchmarks}

We evaluate on six public video benchmarks spanning temporal reasoning, spatial--temporal understanding, expert-domain reasoning, and general-purpose video QA. We follow each benchmark's official protocol unless noted otherwise.

\paragraph{Video-Holmes~\citep{videoholmes_2025}.} A reasoning-centric benchmark comprising $1{,}837$ questions over $270$ narrative videos, where each correct answer requires multi-step inference across non-adjacent clips (e.g., motive attribution, causal event linking, and hidden information recovery). It mainly evaluates models’ temporal and causal reasoning capabilities.

\paragraph{VSI-Bench~\citep{vsibench_2025}.} A spatial-reasoning benchmark for MLLMs built on $288$ real-world room-tour videos and roughly $5{,}000$ question--answer pairs. Its eight tasks (object count, absolute and relative distances, relative direction, route planning, object size, room size, and appearance order) span three reasoning families: configurational, measurement estimation, and spatiotemporal. Each task requires aggregating evidence across many frames rather than from any single salient view.

\paragraph{VideoMMMU~\citep{videommmu_2025}.} An expert-domain benchmark with $900$ questions over $300$ college-level instructional videos covering six disciplines (Art, Business, Science, Medicine, Humanities, Engineering). The questions are organized along three cognitive stages: perception, comprehension, and adaptation, which mirror a human-like progression from observation to applied reasoning. Strong performance therefore demands both the domain knowledge to interpret the demonstration and the temporal grounding that aligns each step of the demonstration to the question being asked.

\paragraph{MMVU~\citep{mmvu_2025}.} A multi-discipline video benchmark of $3{,}000$ expert-annotated question--answer pairs over $1{,}529$ specialized-domain videos that span $27$ subjects across four core disciplines (Science, Healthcare, Humanities \& Social Sciences, Engineering). Following prior work~\cite{feng2025video, park2025deepvideo, wang2026video}, we evaluate on the multiple-choice subset to keep our numbers directly comparable with the published baselines.

\paragraph{TempCompass~\citep{tempcompass_2024}.} A diagnostic benchmark for fine-grained temporal perception, structured along five basic temporal axes: action, speed, direction, attribute change, and event order, and ten finer sub-aspects such as relative speed, camera direction, and combined change.

\paragraph{VideoMME~\citep{videomme_2024}.} A general-purpose video-understanding benchmark of $2{,}700$ questions over $900$ videos drawn from six domains and $30$ sub-class video types. Each video is bucketed by duration into a short (${<}2$~min), medium ($4$--$15$~min), or long ($30$--$60$~min) split. We report the average across the three duration splits \emph{without subtitles}, so the model must rely solely on the visual stream.

\section{Temporal Keyword Frequency}
\label{app:qualitative}
\begin{figure}[h]
    \centering
    \includegraphics[width=\linewidth]{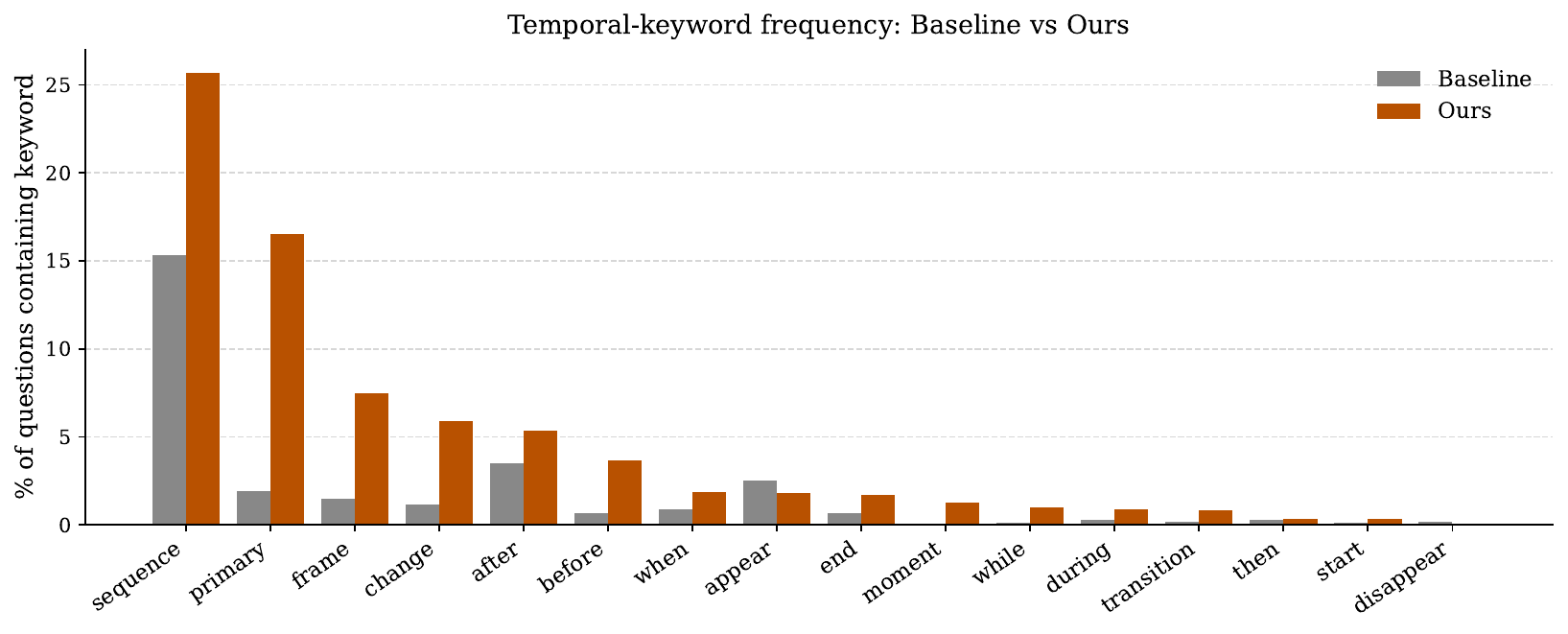}
    \caption{Temporal keyword frequency comparison.}
    \label{fig:freq}
\end{figure}

The word cloud analysis in \S\ref{sec:exp} (Figure~\ref{fig:wordcloud}) shows a lexical shift in the Questioner's vocabulary toward temporal and reasoning words. Figure~\ref{fig:freq} additional presents the frequency change for temporally grounded vocabularies between baseline and \ours. Compared to the baseline, our model significantly increases the usage of key temporal and causal terms such as \emph{sequence}, \emph{primary}, \emph{change}, \emph{after}, and \emph{before}. Notably, high-level reasoning cues (\eg, \emph{sequence}, \emph{change}) exhibit the largest gains, indicating stronger emphasis on temporal progression and event dynamics. The broader coverage of temporal keywords, including less frequent terms like \emph{transition}, \emph{during}, and \emph{while}, further suggests improved diversity in temporal reasoning patterns. Overall, this shift validates that our approach encourages more temporally aware and causally grounded question generation.

\section{Limitations and Broader Impacts}
\label{app:limitations}

\subsection{Limitations}
Despite the promising performance of our video-based self-evolving framework, several limitations remain. First, the current framework is trained with only 16 video frames, which may restrict its ability to model long-range temporal dependencies. Extending the paradigm to handle longer videos and more complex real-world video streams can be potential directions. In addition, our method is primarily motivated by intrinsic supervision signals centered on temporal sensitivity and temporal grounding. Future work could further explore other important aspects of video reasoning, including fine-grained spatial understanding and multi-event dependency modeling.

\subsection{Broader Impacts}
This work presents a video-centric self-evolving framework that enables Video-LLMs to improve temporal reasoning directly from raw, unannotated videos without relying on large-scale human supervision. Our work focuses primarily on methodological research and benchmark evaluation, and does not introduce new negative social impacts. However, it may inherit or amplify biases present in large-scale internet video data or pretrained models, potentially leading to unfair or unreliable behavior across different environments and populations.

\clearpage


\end{document}